\documentclass[conference]{IEEEtran}

\usepackage{cite}
\usepackage{amsmath,amssymb,amsfonts}
\usepackage{algorithmic}
\usepackage{graphicx}
\usepackage{textcomp}
\usepackage[table]{xcolor}
\usepackage{booktabs}
\usepackage{threeparttable}
\usepackage{tabularx}
\usepackage{listings}
\usepackage{url}
\usepackage{bbm}
\usepackage{mdframed}
\usepackage{xspace}

\DeclareMathOperator*{\argmax}{argmax} 

\lstdefinestyle{prompt}{
basicstyle=\ttfamily\scriptsize,
breaklines=true,
breakatwhitespace=true,
columns=fullflexible,
keepspaces=true,
showstringspaces=false,
literate={—}{{---}}1,
aboveskip=4pt,
belowskip=4pt
}

\newcommand\boxedname{Prompt\xspace} 
\newcounter{boxedcounter}
\definecolor{backgroundcolor}{HTML}{F6F3EB}
\newenvironment{pmdframed}[1][]{
    \refstepcounter{boxedcounter}
    \begin{mdframed}[
        innertopmargin=2pt, 
        innerbottommargin=2pt,
        innerleftmargin=2pt, 
        innerrightmargin=2pt,
        frametitle={\textbf{\boxedname~\theboxedcounter:} #1},
        frametitlefont=\footnotesize,
        frametitlerule=true,
        backgroundcolor=backgroundcolor]%
    \setlength{\parindent}{0pt}%
    \setlength{\parskip}{.5em}%
}
{%
    \par%
    \end{mdframed}%
}

\begin{document}

\title{Explainable Suicide Risk Assessment on Social Media with Multi-Task QLoRA}

\author{
\IEEEauthorblockN{Xuan Zhong Feng\IEEEauthorrefmark{1}, Geoffrey Martin\IEEEauthorrefmark{2}, Hexin Dong\IEEEauthorrefmark{1}, Yifan Peng\IEEEauthorrefmark{1}\IEEEauthorrefmark{2}}
\IEEEauthorblockA{\IEEEauthorrefmark{1}\textit{Department of Population Health Sciences},
\textit{Weill Cornell Medicine},
New York, NY, USA}
\IEEEauthorblockA{\IEEEauthorrefmark{2}\textit{Systems Engineering},
\textit{Cornell University},
New York, NY, USA \\
\{xuf4001, ghm4002, hed4006, yip4002\}@med.cornell.edu}
}

\maketitle

\begin{abstract}
Explainable suicide-risk assessment requires models not only to estimate risk severity, but also to identify supporting language and the risk and protective factors expressed in a post. We present our system for the IEEE BigData 2026 Cup on Explainable Suicide Risk Assessment on Social Media, which addresses three tasks: risk-level classification, evidence phrase extraction, and multi-label factor identification. Our approach adapts Qwen2.5-Instruct models using quantized low-rank adaptation (QLoRA) and an answer-masked causal language-model objective. We jointly train across all three tasks for risk classification, jointly train on Tasks~1a and 1b for evidence extraction, and adapt Task~2 separately for factor identification. We also tailor aggregation to each output: we average risk-level probabilities from the 32B and 72B models, combine evidence phrases through cross-fold consensus, and calibrate factor-specific decisions through rate matching based on out-of-fold operating points. On the official leaderboard, the final system achieved a composite score of 0.7738, with 0.8089 on Task~1 and 0.6919 on Task~2. Across the evaluated configurations, three-task training performed best for Task~1a, joint training on Tasks~1a and 1b performed best for Task~1b, and task-specific training performed best for Task~2. Probability averaging further improved Task~1a when component models had complementary errors. These findings highlight the value of tailoring both training objectives and aggregation strategies to the output structure of each task within a unified language-model framework.

\end{abstract}


\section{Introduction}

Suicide is a major public health concern, and timely recognition of suicidal thoughts and behaviors remains an important component of prevention~\cite{turecki2019suicide, horowitz2009screening}. Social-media platforms contain large volumes of spontaneous language about distress, interpersonal conflict, self-harm, and help-seeking that may complement information available through traditional care settings~\cite{gaur2021suicidality, notredame2018socialmedia}. This has motivated the development of automated suicide-risk assessment models from user-generated text. However, because these systems operate in a high-stakes setting, predicting a risk category alone is insufficient. Clinicians and other stakeholders also need to understand the evidence underlying the prediction, including which text supports the assessment and which risk and protective factors are present.

These requirements naturally correspond to multiple natural language processing (NLP) tasks. Suicide-risk assessment involves text classification to determine suicide risk, extractive generation (or rationale extraction) to identify the text spans supporting the prediction, and multi-label classification to recognize the risk and protective factors expressed in a post.  Although these tasks are closely related and operate over the same text input, they differ in their prediction objectives and output structures. They are further complicated by informal language, limited annotated data, and severe class imbalance. 

To advance explainable suicide-risk assessment, the IEEE BigData 2026 Cup on Explainable Suicide Risk Assessment on Social Media formalizes this problem into three complementary subtasks~\cite{ieeebigdata2026cup}. Task~1a assigns a suicide-risk level to each Reddit post, Task~1b extracts the phrases that support that assessment, and Task~2 identifies applicable risk and protective factors. 
In this report, we describe our submission to the shared task. We address the three subtasks with a definition-aware large language model (LLM) framework that represents each task through natural language instructions and structured outputs. This design allows the tasks to share relevant information while preserving their distinct prediction objectives. Because the tasks differ in the density and structure of their supervision, we compared single-task and joint-training configurations and selected the strongest configuration for each output.

Our system also uses task-specific aggregation strategies that reflect the structure of each output. Risk-level predictions are combined in a shared probability space, evidence phrases are aggregated through consensus, and factor decisions use label-specific operating points estimated from out-of-fold predictions. On the official leaderboard, our final system achieved a composite score of 0.7738, with 0.8089 on Task~1 and 0.6919 on Task~2.

The remainder of this paper is organized as follows. We first introduce the shared task and dataset (Section~\ref{sec:task}) and describe the proposed methodology (Section~\ref{sec:method}). We then present the experimental setup (Section~\ref{sec:experiments}) and report the results (Section~\ref{sec:results}), followed by a discussion of the main findings and limitations (Section~\ref{sec:discussion}).

\section{Task and Dataset}
\label{sec:task}

\subsection{Task Definition}

The IEEE BigData 2026 Cup on Explainable Suicide Risk Assessment on Social Media asks systems to assess suicide risk from Reddit posts while also providing interpretable supporting information~\cite{ieeebigdata2026cup}. The shared task comprises three prediction subtasks. 

Task~1a assigns each post one of four risk levels: \emph{Indicator}, indicating no explicit mention of suicide; \emph{Ideation}, indicating explicit suicidal expression without a plan; \emph{Behavior}, indicating suicidal expression accompanied by self-harm or a suicidal plan; and \emph{Attempt}, indicating a recent or past suicide attempt. Task~1b extracts phrases or short clauses from the original post that support the Task~1a judgment. Task~2 identifies all applicable suicide-related factors from a taxonomy of 24 labels.

Task~1a is evaluated with weighted F1 ($F_{\mathrm{risk}}$), 
Task~1b is evaluated with Phrase F1 ($F_{\mathrm{phrase}}$),
and Task~2 is evaluated with macro F1 over the 24 factors ($F_{\mathrm{factor}}$). The overall composite score is
\begin{equation}
S = 0.4\;F_{\mathrm{risk}} + 0.3\;F_{\mathrm{phrase}}
    + 0.3\;F_{\mathrm{factor}},
\label{eq:composite}
\end{equation}

\subsection{Dataset and Annotation}

The competition dataset consists of anonymized posts collected from the \texttt{r/SuicideWatch} community on Reddit. Trained annotators labeled the posts according to suicide-risk definitions grounded in the Columbia-Suicide Severity Rating Scale~\cite{posner2011cssrs}. Each labeled post contains a risk level (Task~1a), zero or more supporting evidence phrases (Task~1b), and zero or more suicide-related factors drawn from a taxonomy of 24 factors (Task~2).

As shown in Table~\ref{tab:data}, the labeled set contains 1,635 posts from 153 users, while the leaderboard set contains 378 unlabeled posts from 36 users. The risk distribution in Task~1a is highly imbalanced: 611 posts (37.4\%) are labeled Indicator, 519 (31.7\%) Ideation, 391 (23.9\%) Behavior, and 114 (7.0\%) Attempt. The annotations contain 1,833 evidence phrases, averaging 1.12 phrases per post, while 588 posts (36.0\%) have no evidence phrase. They also contain 4,772 factor assignments, averaging 2.92 factors per post, while 128 posts (7.8\%) have no factor label. The number of positive examples per factor ranges from 8 to 745 posts, also indicating substantial class imbalance.

\begin{table}[t]
\caption{Training-set distributions across the three shared tasks.}
\label{tab:data}
\centering
\begin{tabular}{lrr}
\toprule
& \textbf{\(n\)} & \textbf{\%} \\
\midrule
\rowcolor{gray!30}Task~1a: Risk level & &\\
Indicator & 611 & 37.4 \\
Ideation & 519 & 31.7 \\
Behavior & 391 & 23.9 \\
Attempt & 114 & 7.0 \\
\midrule
\rowcolor{gray!30}Task~1b: Evidence phrases & & \\
0 & 588 & 36.0 \\
1 & 587 & 35.9 \\
2 & 258 & 15.8 \\
$\geq 3$ & 202 & 12.4 \\
\midrule
\rowcolor{gray!30}Task~2: Factors &&\\
Cognitive deficits & 33 & 2.0 \\
Coping strategy & 443 & 27.1 \\
Dysfunctional family & 118 & 7.2 \\
Emotion dysregulation & 542 & 33.1 \\
Exposure to others' suicide & 14 & 0.9 \\
Hopelessness & 745 & 45.6 \\
Interpersonal difficulty & 195 & 11.9 \\
Interpersonal violence & 83 & 5.1 \\
Low self-esteem & 475 & 29.1 \\
Low socio-economic status & 54 & 3.3 \\
Meaning in life & 45 & 2.8 \\
Mental health issues & 228 & 13.9 \\
Physical health/characteristic & 78 & 4.8 \\
Poor school performance & 16 & 1.0 \\
Poor social support & 501 & 30.6 \\
Prior self-harm or suicidal thought/attempt & 215 & 13.1 \\
Psychological capital & 238 & 14.6 \\
Sense of responsibility & 58 & 3.5 \\
Sexual orientation related issues & 8 & 0.5 \\
Social support & 112 & 6.9 \\
Stressful life event & 182 & 11.1 \\
Substance use & 33 & 2.0 \\
Suicide means (with access) & 292 & 17.9 \\
Traumatic experience & 64 & 3.9 \\
\bottomrule
\end{tabular}
\end{table}

\section{Methodology}
\label{sec:method}

\subsection{System Overview}

\begin{figure*}[!t]
    \centering
    \includegraphics[width=\textwidth]{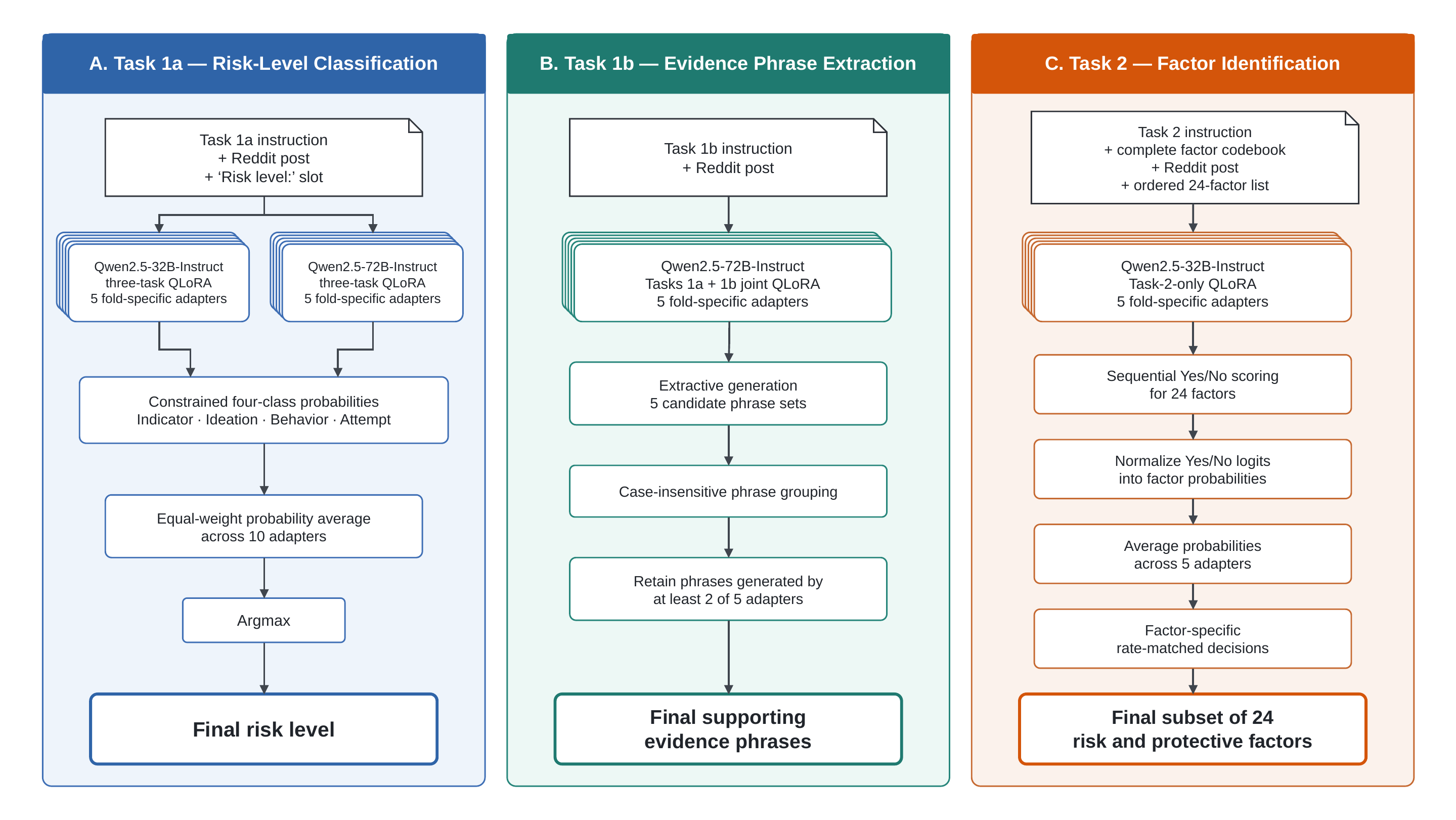}
    \caption{System architecture for the three prediction tasks. Task~1a applies cross-model probability averaging. Task~1b uses cross-fold phrase consensus. Task~2 makes factor-level decisions using rate-matched operating points estimated from out-of-fold predictions.}
    \label{fig:system-architecture}
\end{figure*}

Our system uses a common instruction-based LLM interface across all three tasks, while applying task-specific inference and aggregation procedures to accommodate their different output structures (Fig.~\ref{fig:system-architecture}). The final system contains four sets of five fold-specific models. 
First, we jointly trained Qwen2.5-32B-Instruct and Qwen2.5-72B-Instruct on all three tasks~\cite{yang2024qwen25}. These models contribute to risk-level classification. 
Second, we trained Qwen2.5-72B-Instruct on Task~1a and Task~1b to generate evidence spans. 
Third, we trained a separate Qwen2.5-32B-Instruct exclusively on Task~2 to produce the final suicide-factor predictions.

Thus, the system uses different aggregation strategies for the three subtasks. For Task~1a, it applies cross-model probability averaging. For Task~1b, it uses cross-fold phrase consensus. For Task~2, it makes factor-level decisions using rate-matched operating points estimated from out-of-fold predictions.

\subsection{Definition-Aware Multi-Task Adaptation}

Across all three tasks, we use a common definition-aware adaptation framework with task-specific instructions and output formats. Specifically, we adapt instruction-tuned Qwen2.5 models~\cite{yang2024qwen25} using quantized low-rank adaptation (QLoRA)~\cite{dettmers2023qlora}. For each post \(x\) and task \(t\), we construct a task-specific instruction \(q_t(x)\) and target response \(a_t\). The complete task-specific templates are provided in Appendix~\ref{app:prompts}.

During training, we mask prompt tokens and fixed output scaffolding, such that the causal language-model loss is computed only over answer tokens:
\begin{equation}
\mathcal{L}_t(x) =
-\frac{1}{|A_t|}
\sum_{i\in A_t}
\log p_\theta\!\left(a_i \mid q_t(x),a_{<i}\right),
\label{eq:masked-loss}
\end{equation}
where \(A_t\) is the set of answer-token positions. At each answer position $i$, the model produces a vocabulary-sized logit vector $z_i$, from which the probability of the target next token $a_i$ is obtained by
\begin{equation}
p_\theta(a_i \mid q_t(x),a_{<i})= \operatorname{softmax}(z_i)_{a_i}.
\end{equation}

Thus, optimization is restricted to the tokens that constitute the task-specific answer, while prompt tokens and fixed output scaffolding do not contribute to the loss. Under QLoRA, the pretrained model parameters remain frozen while low-rank adapter parameters are optimized, allowing task-specific adaptation while preserving the pretrained language-model output space.
Under joint training, each post yields one training example for each included task, enabling the model to learn a shared representation.

\subsection{Risk-Level Classification (Task~1a)}

We formulate Task~1a as a forced-slot classification task. Following the instruction and post, the prompt ends with a fixed ``\texttt{Risk level:}'' prefix (Prompt~\ref{prompt:task1a}). The model is trained using cross-entropy loss computed at the answer position, where the target corresponds to the correct risk label $y\in Y :=\{\text{Indicator}, \text{Ideation}, \text{Behavior}, \text{Attempt}\}$.

Because the four labels are unique, we treat the four candidate risk labels as a constrained classification space. 
The final risk label is then the class with the highest averaged probability. 
For model $m$, let $z_{m,y}$ be the next-token logit assigned to the first token of risk label $y$ immediately after the fixed ``\texttt{Risk level:}'' prefix. We convert these four label logits into normalized class probabilities and ensemble them across models:
\begin{equation}
\hat{y} = \argmax_{y\in Y} \frac{1}{|M|}\sum_{m \in M}
\frac{\exp z_{m,y}}
{\sum_{y'\in Y}\exp z_{m,y'}}.
\label{eq:t1a-prob}
\end{equation}
This design maps the next-token predictions to a common four-class probability space and supports direct comparison and equal-weight averaging across models.
The final Task~1a ensemble contains five fold-specific 32B adapters and five fold-specific 72B adapters, with one adapter of each model size trained for each cross-validation fold. For each post, we obtain the four-class probability distribution from each of the ten adapters, average the probabilities with equal weights, and select the class with the highest average probability.

For ablation studies, we trained adapters on Task~1a alone, Tasks~\{1a, 1b\} jointly, or all three tasks, using the same optimizer hyperparameters and number of epochs. We also included an encoder-based baseline by fine-tuning MentalRoBERTa~\cite{ji2022mentalbert} with a class-weighted four-class softmax classification head and a 512-token input limit. In each outer fold, we used three folds for training, one for checkpoint selection, and one for evaluation. We trained the models for 10 epochs using AdamW with a batch size of 16, a learning rate of $2\times10^{-5}$, weight decay of 0.01, and random seed 42.

\subsection{Evidence Phrase Extraction (Task~1b)}

Because Task~1b requires evidence spans from the original post, we formulate it as an extractive generation task.
Therefore, the instruction asks the model to copy the shortest spans that justify the assigned risk level and preserve the original wording exactly (Prompt~\ref{prompt:task1b}). 

Each of the five fold-specific adapters independently generates a set of candidate phrases. We then group them case-insensitively and retain only those produced by at least two of the five adapters. We selected this 2-of-5 consensus threshold based on the resulting output density. Taking the union of all five adapters produced an average of 1.521 phrases per post, whereas requiring agreement from two adapters reduced this to 1.146 phrases per post, closer to the out-of-fold (OOF) rate of 0.985 phrases per post. Similar to the Task~1a ablation, we trained Task~1b-only adapters, jointly trained Tasks~\{1a, 1b\} adapters, and three-task adapters using the same hyperparameters.

\subsection{Factor Identification (Task~2)}

For Task~2, the prompt presents the complete factor codebook, followed by the post and an ordered list of the 24 factor names (Prompt~\ref{prompt:task2}). The target response for each factor is a single token, \texttt{Yes} or \texttt{No}. The loss is computed only at these 24 answer positions, encouraging the model to adapt its pretrained language representations to the factor definitions and annotation scheme while maintaining a shared autoregressive interface.

At inference, the model first processes the shared instruction, factor definitions, and post. It then evaluates the 24 factors sequentially, reusing the key--value cache to avoid recomputing the shared prefix. For each factor \(f\), let $z^{\mathrm{Yes}}{m,f}$ and $z^{\mathrm{No}}{m,f}$ denote the logits for the \texttt{Yes} and \texttt{No} tokens, respectively, at the corresponding answer position for model $m$. We compute the probability of a positive response by normalizing these two logits:
\begin{equation}
p_{f}(x)= \frac{1}{|M|}\sum_{m\in M}
\frac{\exp z^{\mathrm{Yes}}_{m,f}}
{\exp z^{\mathrm{Yes}}_{m,f}+\exp z^{\mathrm{No}}_{m,f}}.
\label{eq:t2-prob}
\end{equation}
After selecting the answer for factor $f$, we appended the predicted factor to the context before evaluating the next factor. 
Thus, predictions for later factors can condition on earlier factor decisions.

We estimate a separate operating point for each factor using pooled out-of-fold (OOF) predictions. For each factor \(f\), we first choose the OOF threshold  $\tau_f^*$  that maximizes binary F1. We then apply this threshold $\tau_f^*$ to all posts in the pooled OOF set and calculate the positive prediction rate $r_f$.  Finally, we apply the quantile-based cutoff to the ensembled test probabilities:
\begin{align}
\tau_f^* &=\arg\max_{\tau\in\mathcal{T}}
\;\text{F1}_{f}\!\left(
y_f,\mathbbm{1}[p_f^{\mathrm{OOF}}\geq\tau]
\right),\\
r_f & =\frac{1}{N}\sum_{i=1}^{N}
\mathbbm{1}\!\left[p_{if}^{\mathrm{OOF}}\geq\tau_f^*\right],\\
\hat y_{if}&=\mathbbm{1}
\!\left[\bar p_f(x_i)\geq Q_{1-r_f}(\bar p_{\cdot f})\right],
\label{eq:t2-rate-match}
\end{align}

We do not apply the raw OOF threshold directly to the test set because probability averaging changes the score scale. 
This rate-matching step preserves each factor's OOF operating point while adapting the threshold to the probability distribution of the ensemble.

For ablation studies, we trained adapters on different base LLMs and fine-tuned MentalRoBERTa with a 24-output linear head applied to the pooled classification-token representation. We trained the encoder baseline with an independent factor-wise asymmetric loss, using $\gamma_{-}=2$, $\gamma_{+}=1$, and probability clipping of 0.05. Models used a 512-token input limit, a learning rate of $3\times10^{-5}$, an effective batch size of 16, and at most 12 epochs. Checkpoint selection and factor-specific threshold estimation were performed using an inner validation fold.

We also evaluated per-factor late fusion using GPT-5.4~\cite{openai2026gpt54}. Each post was processed independently with three prompt families based on definitions and coding guidance from the National Violent Death Reporting System (NVDRS) Web Coding Manual~\cite{cdc2025nvdrs}. The prompt families incorporated different levels of coding guidance and factor-specific strictness, including a concise NVDRS-essence prompt containing only the factor definition, an NVDRS-recommended prompt with longer coding guidance for selected complex factors, and a strictness-control prompt with factor-specific evidence thresholds (Appendix~\ref{app:prompts}, Prompts~\ref{prompt:gpt54-essence}--\ref{prompt:gpt54-strict}). For each factor, we selected the prompt family with the highest factor-level F1 across the five validation folds and used that family's binary prediction for the final decision.


\section{Experimental Setup}
\label{sec:experiments}

\subsection{Training Configuration}


We used Qwen2.5-32B-Instruct and Qwen2.5-72B-Instruct~\cite{yang2024qwen25} as the backbone LLMs. For Task~2 backbone ablation, we additionally evaluated Qwen3-32B and Llama-3.3-70B-Instruct~\cite{qwen3,llama33}. Base-model weights were loaded with 4-bit NormalFloat quantization, double quantization, and bfloat16 computation. LoRA updates were applied to the query, key, value, attention-output, gating, up-projection, and down-projection modules. We used gradient checkpointing to reduce memory consumption. Each model was trained on one NVIDIA A100-SXM4-80GB GPU.


For optimization, we used AdamW with zero weight decay, a learning rate of \(1\times10^{-4}\), a batch size of 1 with gradient accumulation of 8, and 2 training epochs. We used 3\% warmup followed by cosine decay. The maximum post-token budget was 1,024, and all experiments used a random seed of 42. The effective batch size was therefore 8.

For Task~1b, we used deterministic greedy decoding with a maximum of 128 newly generated tokens. Tasks~1a and 2 did not require free-text decoding. Their class probabilities were obtained directly from the constrained answer positions described in Section~\ref{sec:method}. For the Task~2-only training setup, we additionally clipped the gradient norm at 1.0.

For each Task~2 backbone, the zero-shot control used the base instruction-tuned model without an adapter. Otherwise, it retained the same Task~2 codebook, answer grid, autoregressive Yes/No scoring procedure, and post-token budget (Appendix~\ref{app:prompts}, Prompt~\ref{prompt:task2}).

\subsection{Cross-Validation and Evaluation Protocol}
\label{sec:cv}


Unless otherwise noted, we used five-fold cross-validation, training the model on four folds and evaluating it on the remaining held-out fold. Each table caption specifies the task-specific metric applied to the pooled out-of-fold predictions.

All experiments used the same five user-grouped folds. Because each user can contribute multiple posts, randomly dividing individual posts could leak author-specific language and context between training and validation sets. We therefore grouped posts by user and stratified the folds by risk level using a random seed of 42. The resulting folds contain 277--410 posts and 29--32 users each, with no user appearing in more than one fold. 

For Task~1b, the reported validation scores are calculated from pooled OOF predictions, with each post evaluated by the single adapter that held it out. The final leaderboard system instead applies 2-of-5 consensus across all fold-specific adapters, so the OOF scores do not directly evaluate the final consensus ensemble.

For the model-based Task~2 configurations in Table~\ref{tab:t2-results}, model fitting and factor-threshold or averaging-policy estimation for a held-out fold excluded that fold's labels. We then pooled the resulting held-out predictions across the five folds and scored them once. We evaluated the GPT-5.4 prompt ensemble separately as a descriptive prompt-selection experiment: we chose its fixed factor-to-prompt mapping using aggregate factor-level F1 across the same five validation folds. Its pooled score therefore is not a nested held-out estimate and should not be compared directly with the fully held-out model-based estimates.

\section{Results}
\label{sec:results}

\subsection{Official Leaderboard Performance}

The final leaderboard system used all five fold-specific adapters for each component, with aggregation tailored to each task's output structure. For Task~1a, we averaged class probabilities from five Qwen2.5-32B adapters and five Qwen2.5-72B adapters. For Task~1b, we retained evidence phrases generated by at least two of the five Qwen2.5-72B adapters jointly trained on Tasks~\{1a, 1b\}. For Task~2, we averaged factor probabilities from five Task~2-only Qwen2.5-32B adapters and then applied the factor-specific rate-matched decision rules. The resulting system achieved a composite leaderboard score of 0.7738, with scores of 0.8089 on Task~1 and 0.6919 on Task~2.

\subsection{Task 1a: Risk-Level Classification}

We compared Qwen2.5-32B models trained on Task~1a alone, jointly on Tasks~\{1a, 1b\}, and jointly on all three tasks (Table~\ref{tab:t1a-results}). We also evaluated the three-task Qwen2.5-72B model and cross-model probability ensembles.

For Qwen2.5-32B, the three-task configuration achieved the highest weighted F1 (0.8259), compared with 0.8209 for Task~1a-only training and 0.8205 for joint training on Tasks~\{1a, 1b\}. The three-task Qwen2.5-72B model achieved a weighted F1 of 0.8285. Finally, averaging the class-probability vectors from the three-task 32B and 72B models produced the best overall performance, with a weighted F1 of 0.8395.

\begin{table}[t]
\caption{Results for Task~1a. Weighted F1 on pooled five-fold out-of-fold predictions ($n{=}1635$).}
\label{tab:t1a-results}
\centering
\begin{tabular}{@{}llr@{}}
\toprule
\textbf{Model} & \textbf{Strategy} & \textbf{Weighted F1} \\
\midrule
MentalRoBERTa Classifier & - & $0.7636$ \\
\midrule
Qwen2.5-32B & 1a & $0.8209$ \\
 & 1a + 1b & $0.8205$ \\
 & 1a + 1b + 2 & $0.8259$ \\
\midrule
Qwen2.5-72B & 1a & $0.8272$ \\
 & 1a + 1b & $0.8225$ \\
 & 1a + 1b + 2 & $0.8285$ \\
\midrule
Qwen2.5-32B + Qwen2.5-72B & 1a + 1b + 2 & $\mathbf{0.8395 }$ \\
\bottomrule
\end{tabular}
\end{table}

To find the best ensemble members, we encoded each model’s pooled OOF predictions as a binary error vector (1 for incorrect predictions and 0 for correct predictions). We computed pairwise Pearson correlations across all 1,635 posts. Lower correlation indicates that two models make mistakes on more distinct examples. The same-base 32B variants had error correlations of 0.79--0.83, whereas the joint 32B and 72B models had a lower correlation of 0.68. These final models disagreed on 158 posts. On this subset, the 32B and 72B models achieved accuracies of 0.462 and 0.487, respectively, while their soft-voting ensemble achieved an accuracy of 0.601.

As an encoder-only baseline, we evaluated a four-class MentalRoBERTa~\cite{ji2022mentalbert} classifier across the same five held-out folds. It achieved a pooled OOF weighted F1 of 0.7636.

\subsection{Task 1b: Evidence Phrase Extraction}

We compared Qwen2.5-32B and Qwen2.5-72B models trained on Task~1b alone, jointly on Tasks~\{1a, 1b\}, or jointly on all three tasks, together with a MentalRoBERTa+CRF baseline. Table~\ref{tab:t1b-results} summarizes these results.


\begin{table}[t]
\caption{Results for Task~1b. Phrase F1, defined as the mean post-level score, on pooled five-fold out-of-fold predictions ($n{=}1635$).}
\label{tab:t1b-results}
\centering
\begin{tabular}{@{}llr@{}}
\toprule
\textbf{Model} & \textbf{Strategy} & \textbf{Phrase F1}\\
\midrule
MentalRoBERTa + CRF & -- & $0.7044$ \\
\midrule
Qwen2.5-32B & 1b & $0.7676$ \\
 & 1a + 1b & $0.7792$ \\
 & 1a + 1b + 2 & $0.7821$ \\
\midrule
Qwen2.5-72B & 1b & $0.7761$ \\
  & 1a + 1b & $\mathbf{0.7874}$ \\
 & 1a + 1b + 2 & $0.7786$ \\
\bottomrule
\end{tabular}
\end{table}

For Qwen2.5-32B, Phrase F1 increased from 0.7676 with Task~1b-only training to 0.7792 with joint training on Tasks~\{1a, 1b\} and 0.7821 with joint training on all three tasks. For Qwen2.5-72B, joint training on Tasks~\{1a, 1b\} achieved the highest single-adapter pooled OOF Phrase F1 of 0.7874, compared with 0.7761 for Task~1b-only training and 0.7786 for three-task training. We therefore selected the Qwen2.5-72B adapters jointly trained on Tasks~\{1a, 1b\} for the final system. As described in Section~\ref{sec:cv}, these OOF scores are based on one held-out adapter per post rather than the final 2-of-5 cross-fold consensus.

\subsection{Task 2: Suicide-Factor Identification}

We compared a MentalRoBERTa encoder baseline, the GPT-5.4 per-factor prompt ensemble, full and parameter-efficient LLM adaptation, and cross-model probability ensembles. Table~\ref{tab:t2-results} reports five-fold held-out results using fold-specific model fitting and decision-rule estimation, except for the GPT-5.4 prompt ensemble.

\begin{table}[t]
\caption{Pooled five-fold Task~2 results over $n{=}1635$ posts.}
\label{tab:t2-results}
\centering
\begin{threeparttable}
\begin{tabular}{@{}llr@{}}
\toprule
\textbf{Model} & \textbf{Strategy} & \textbf{Macro-F1} \\
\midrule
MentalRoBERTa & Asymmetric Loss & $0.4507$ \\
\midrule
Zero-shot & Qwen2.5-32B & $0.6384$ \\
 & Qwen2.5-72B & $0.6507$ \\
 & Qwen3-32B & $0.6345$ \\
 & Llama-3.3-70B & $0.6225$ \\
 & GPT-5.4\tnote{*} & $0.6191$ \\
\midrule
Task~2-only QLoRA & Qwen2.5-32B & $0.7000$ \\
 & Qwen2.5-72B & $0.7207$ \\
 & Qwen3-32B & $0.7036$ \\
 & Llama-3.3-70B & $0.6941$ \\
 & Qwen3-32B + Qwen2.5-32B & $0.7160$\\
 & Qwen2.5-32B + Qwen2.5-72B & $\mathbf{0.7309}$ \\
\midrule
Three-task QLoRA & Qwen2.5-32B & $0.6683$ \\
& Qwen2.5-72B & $0.6847$\\
\bottomrule
\end{tabular}
\begin{tablenotes}[flushleft]
\item[*] Descriptive, non-nested estimate: the factor-to-prompt mapping was selected using aggregate factor-level F1 across the same five folds.
\end{tablenotes}
\end{threeparttable}
\end{table}

MentalRoBERTa achieved a macro-F1 of 0.4507, and the GPT-5.4 per-factor prompt ensemble achieved 0.6191, both lower than the adapted LLMs. The adapter-free zero-shot controls achieved macro F1 scores of 0.6384 with Qwen2.5-32B, 0.6507 with Qwen2.5-72B, 0.6345 with Qwen3-32B, and 0.6225 with Llama-3.3-70B. 

QLoRA improved performance over the corresponding zero-shot baselines for all four backbones, with gains ranging from 0.0616--0.0716. Under the same nested decision-rule protocol, QLoRA with Qwen2.5-32B achieved a macro F1 of 0.7000, compared with 0.6734 for full fine-tuning. Three-task QLoRA achieved 0.6683 with Qwen2.5-32B and 0.6847 with Qwen2.5-72B, both below the corresponding Task~2-only QLoRA models, which achieved 0.7000 and 0.7207, respectively.

Cross-model ensembling further improved performance. 
Averaging the probabilities from Qwen3-32B and Qwen2.5-32B reached a macro F1 of 0.7160, while averaging Qwen2.5-32B and Qwen2.5-72B produced the best OOF score of 0.7309. Despite this OOF advantage, the final submission retained Qwen2.5-32B alone because its leaderboard score of 0.6919 was slightly higher than that of Qwen2.5-72B (0.6912).

For the Task~2-only Qwen2.5-32B QLoRA model, mean per-factor OOF AUROC was approximately 0.96, whereas macro-F1 under the nested decision-rule protocol was 0.7000. This difference reflects the distinct properties of the two metrics. AUROC measures how well the model ranks positive examples above negative ones independently of a decision threshold, while macro-F1 evaluates the resulting binary predictions after thresholding. The contrast suggests that the model separated positive and negative examples well, but converting those scores into reliable decisions remained difficult, particularly for factors with few positive examples.

\subsection{Cross-Task Findings}

The optimal training-task mixture depended on both the target task and the model backbone. For Task~1a, three-task training achieved the strongest performance for both backbones: weighted F1 scores of 0.8259 for Qwen2.5-32B (compared with 0.8209 for Task~1a-only and 0.8205 for Tasks~\{1a,1b\}), and 0.8285 for Qwen2.5-72B (compared with 0.8272 and 0.8225, respectively). For Task~1b, joint training improved the 32B model, with Phrase F1 increasing from 0.7676 with Task~1b alone to 0.7792 with Tasks~\{1a,1b\} and 0.7821 with all three tasks. In contrast, the 72B model achieved its best performance with Tasks~\{1a,1b\}, reaching 0.7874 compared with 0.7761 for Task~1b-only and 0.7786 for three-task training. Task~2 performed best with a dedicated adapter among the tested configurations. We therefore selected the training-task mixture independently for each output.

The experiments also show that the ensemble mechanism should match the structure of the output space. For Task~1a, probability averaging was effective when combining models with complementary errors. Task~2 likewise benefited from factor-wise probability averaging in a shared 24-dimensional output space. However, improvements in OOF performance did not always translate to stronger leaderboard performance, likely because macro-F1 was sensitive to threshold selection and rare factors. Together, these observations motivated an output-specific strategy in which training objectives and aggregation procedures are tailored to the characteristics of each task.

\section{Discussion}
\label{sec:discussion}


Three findings shaped the final system. First, the benefit of auxiliary supervision is associated with the density and complexity of the target. Task~1a, with just one class target per post, benefited from the richer Task~1b and Task~2 supervision. For Task~1b, adding Task~1a supervision improved Phrase F1 at both model sizes; adding Task~2 further improved the 32B model but reduced Phrase F1 for the 72B model. Task~2, with 24 decisions per post, performed best with dedicated adaptation. This pattern suggests that auxiliary supervision can help lower-density targets until competing supervision begins to interfere. However, because joint settings processed more examples and optimizer updates, these experiments
are not enough to establish a causal effect of auxiliary supervision or label density.

Second, ensemble effectiveness depended on both the compatibility of the output representation and the complementarity of the component predictions. For Task~1a, the 32B and 72B models expressed their decisions in the same four-dimensional probability space and made complementary errors, allowing probability averaging to increase weighted F1 to 0.8395. Task~2 similarly represented predictions as aligned probabilities for 24 factors; averaging the Qwen2.5-32B and Qwen2.5-72B outputs increased nested macro-F1 from 0.7000 and 0.7207, respectively, to 0.7309. Thus, a shared prediction space enables aggregation, while performance gains depend on the component models contributing complementary information.

Third, the highest OOF score did not always identify the best leaderboard system. This was most visible for Task~2, where larger models and cross-model ensembles improved pooled OOF macro-F1 but did not surpass Qwen2.5-32B on the leaderboard set. Rare factors make both threshold estimation and macro-F1 highly variable, so small aggregate validation gains should be interpreted cautiously. In this setting, the simpler Qwen2.5-32B QLoRA model also outperformed full fine-tuning, suggesting that parameter-efficient adaptation was not only more economical but better regularized for the available training set. However, this comparison is specific to our training configuration and does not establish a general advantage of QLoRA.

This work has four limitations. First, the dataset is small and drawn from a single Reddit community, with 153 labeled users, only 114 Attempt posts, and as few as eight positive examples for some factors. Therefore, the results may not generalize to other platforms, populations, time periods, or clinical settings. Also, post-level modeling does not capture longitudinal changes in risk. Second, the single-task and joint-training comparisons used the same number of epochs, but the number of optimizer updates was not controlled. Joint configurations therefore processed more task-specific examples and received more optimizer updates, so these comparisons do not isolate auxiliary supervision from additional optimization. Third, we compared QLoRA with full fine-tuning only for Qwen2.5-32B, so we cannot determine whether its observed advantage extends to other model families or scales. Fourth, even with QLoRA, using large language models entails substantial computational cost. Future work should evaluate update-matched task mixtures and repeated training seeds, compare adaptation strategies across additional backbones, improve calibration and threshold estimation for rare factors, and evaluate the system on external and longitudinal data before considering model compression or deployment.

\section{Conclusion}
\label{sec:conclusion}

We presented a definition-aware, instruction-based framework for risk classification, supporting-evidence extraction, and suicide-factor identification. The final system used joint adaptation and probability ensembling for risk classification, cross-fold phrase consensus for evidence extraction, and task-specific adaptation with factor-wise decision rules for multi-label identification. Our experiments show that related supervision can improve tasks with lower-density labels, whereas the more densely supervised factor-identification task benefits from separate adaptation. However, because training exposure differed across configurations, these associations should not be interpreted causally. The results also demonstrate that effective ensemble design should account for both the structure of the prediction space and the predictive complementarity of the component models. Future work should improve estimation for rare factors and evaluate the system on external and longitudinal data. Given the sensitivity of suicide-related predictions, any use beyond research would require clinical validation, strong privacy safeguards, and human oversight.

\bibliographystyle{IEEEtran}
\bibliography{ref}

\newpage

\appendices

\section{Prompt Templates}
\label{app:prompts}

The first three listings reproduce the user instructions and assistant answer formats used by the final Qwen system. The later listings document the three GPT-5.4 prompt families used in the Task~2 prompt-ensemble experiment. The placeholder \texttt{<POST>} was replaced by the post after task-specific preprocessing. Standard chat-template control tokens are omitted because they are model- or API-specific wrappers rather than part of the instruction text. During Qwen training, the bracketed answer placeholders were replaced by the gold outputs; during inference, the model supplied or scored these outputs.

\subsection{Risk-Level Classification (Task~1a)}

\begin{pmdframed}[Risk-Level Classification (Task~1a)]
\label{prompt:task1a}
\begin{lstlisting}[style=prompt]
You are assessing suicide risk in a Reddit post, following a strict codebook.

Risk levels, from lowest to highest:
- Indicator: risk factors are present but the author expresses no suicidal thoughts.
- Ideation: the author expresses suicidal thoughts or a wish to die, without plan or act.
- Behavior: the author describes a plan, preparation, or self-harm without a suicide attempt.
- Attempt: the author describes an actual suicide attempt.

Post:
"""
<POST>
"""

Assign exactly one risk level.
\end{lstlisting}
\end{pmdframed}

After the generation prompt, the following forced answer slot was appended:

\begin{lstlisting}[style=prompt]
Risk level: <Indicator|Ideation|Behavior|Attempt>
\end{lstlisting}

\subsection{Evidence Phrase Extraction (Task~1b)}

\begin{pmdframed}[Evidence Phrase Extraction (Task~1b)]
\label{prompt:task1b}
\begin{lstlisting}[style=prompt]
You are extracting evidence for suicide-risk assessment from a Reddit post.

Quote the exact spans of the post that evidence the author's suicidal risk -- the phrases an annotator would highlight to justify the risk level. Copy each span VERBATIM from the post, one per line, shortest span that carries the evidence. Do not paraphrase, explain, or number them.
Many posts contain no such evidence; for those answer with the single word NONE.

Post:
"""
<POST>
"""

Evidence spans:
\end{lstlisting}
\end{pmdframed}

The assistant returned one verbatim span per line, or \texttt{NONE}:

\begin{lstlisting}[style=prompt]
<EVIDENCE SPAN 1>
<EVIDENCE SPAN 2>
...
\end{lstlisting}

\subsection{Suicide-Factor Identification (Task~2)}

\begin{pmdframed}[Suicide-Factor Identification (Task~2)]
\label{prompt:task2}
\begin{lstlisting}[style=prompt,basicstyle=\ttfamily\tiny]
You are annotating Reddit posts for suicide-risk factors, following a strict codebook. Below are 24 factors with their definitions.

1. mental health issues: The user is experiencing a mental health condition, diagnosis, treatment context, psychiatric hospitalization, or clearly named symptoms such as depression, PTSD, psychosis, panic disorder, eating disorder, or BPD. Do not use this label for ordinary sadness, stress, or suicidality alone.
2. physical health/characteristic: The user is experiencing a physical illness, disability, injury, pain, sleep problem, medication side effect, body-function problem, or distress about a concrete body/health characteristic. Do not use this label for emotional hurt or tiredness unless a physical condition is described.
3. substance use: Uncontrolled, harmful, or salient use of drugs, alcohol, tobacco, or other substances.
4. hopelessness: Hopelessness, pointlessness, meaninglessness, emptiness, feeling trapped, stuck, or unable to escape a cycle.
5. emotion dysregulation: The user is experiencing intense or poorly controlled emotion such as panic, fear, rage, acute distress, emotional pain, spiraling, uncontrollable crying, mood swings, or inability to calm down. Do not replace this with mental health issues unless a condition or treatment is also described.
6. low self-esteem: Feeling like a failure, worthless, ugly, a loser, a burden, inadequate, ashamed, or inferior.
7. poor school performance: The user is experiencing failing grades, falling behind in school, inability to keep up academically, or expulsion. Do not use this label for general school/college mentions or non-academic school stress.
8. low socio-economic status: The user is experiencing poverty, unemployment, job loss, housing insecurity, homelessness, debt, lack of money, being unable to afford needs, or financial survival stress. Do not use this label for general job/school dissatisfaction unless financial hardship is stated.
9. interpersonal violence: Harassment, bullying, abuse, assault, domestic violence, threats, coercion, or verbal/physical violence from another person.
10. prior self-harm or suicidal thought/attempt: The user has a history of self-harm (cutting, burning), a prior suicide attempt at any point in life, prior or recurring suicidal thoughts that predate this post (e.g. 'I've been thinking about this for years', previously told someone they wanted to die), a current act of self-harm, a written suicide note, or explicit current suicidal planning/intent. Do not use this label when the only evidence is first-time suicidal ideation expressed in the current post with no history, behavior, or concrete plan.
11. poor social support: The user is experiencing loneliness, isolation, rejection, abandonment, neglect, lack of friends/support, no one to turn to, or asking strangers for someone to talk to because support is missing. Do not use social support merely because the user requests help.
12. interpersonal difficulty: The user is experiencing relationship conflict, breakup, rejection, social difficulty, inability to connect, communication problems, bullying/rejection without violence, or conflict with friends/partners/peers. Distinguish from poor social support, which is about lack of support or isolation.
13. dysfunctional family: Conflict or problems with family members (parents, siblings, children, relatives — not intimate partners): estrangement, ongoing arguments, neglectful, indifferent, controlling, or unsafe family dynamics, stressors hitting the family unit, or substance abuse in the household. Do not use this label for routine family disagreements or for intimate-partner conflict.
14. exposure to others' suicide: The user is experiencing exposure to another person's suicide, suicide attempt, self-harm, or serious suicidal crisis among friends, family, partners, or close others. Do not use this for the user's own suicidality.
15. stressful life event: Pregnancy, robbery, pandemic-related uncertainty, relationship loss, legal trouble, job/school crisis, bereavement, or other acute stressful events.
16. traumatic experience: Trauma, abuse, neglect, assault, violence exposure, accident, disaster, or other deeply distressing past experience.
17. cognitive deficits: Memory problems, confusion, brain fog, impaired concentration, feeling unable to think clearly, or other cognitive difficulties.
18. suicide means (with access): The user is experiencing access to, preparation for, or discussion of a concrete suicide method such as pills, gun, rope, hanging, jumping, vehicle, bridge, overdose, blade, train, poison, drowning, or other means. Do not use this for general desire to die or vague references to killing oneself without a method/access.
19. sexual orientation related issues: Gender/sexual disorder, sexual orientation-related distress, same-sex relationship issues, identity conflict, stigma, or related suicidal thoughts.
20. social support: PROTECTIVE: supportive friends, family, partners, professionals, or community are present and actively providing or offering meaningful help (e.g. 'my friends have been checking on me', 'my family is supporting me through this'). Do not use this label merely because the user is asking strangers online for support, or when support is absent, strained, or conditional.
21. coping strategy: The user describes a deliberate effort to cope with or manage distress: therapy, journaling, exercise, art, mindfulness, talking to someone, calling a hotline, self-soothing, distracting, resting to stay safe, trying to stay positive, or posting here to seek help or vent — reaching out for help is itself a coping strategy. Do not count suicide planning as coping unless the action reduces harm or seeks help.
22. psychological capital: The user is experiencing hope, resilience, agency, desire to improve, motivation to keep going, self-efficacy, wanting happiness, trying, courage, or belief that things could get better. Code this even when distress is severe; do not require the post to be mostly positive.
23. sense of responsibility: The user is experiencing duty, obligation, commitment, guilt about impact on others, responsibility to children/family/pets/work/school, or desire not to hurt/burden others. Do not use this for general guilt unless a concrete responsibility or concern for others is present.
24. meaning in life: PROTECTIVE: the user cites a purpose, goal, faith, or concrete reason for living that they currently hold onto (e.g. 'my kids keep me going', 'my dog needs me', 'I still have things I want to accomplish'). Do not use this label for despairing rhetorical statements such as 'what's the point' or for the absence of meaning — those indicate hopelessness instead.

Post:
"""
<POST>
"""

For each of the 24 factors, answer Yes only if the post gives clear evidence for it under the definition above; otherwise answer No. Answer every factor in order.
\end{lstlisting}
\end{pmdframed}

The following ordered answer grid was appended to the assistant turn. Each placeholder was represented by a single \texttt{Yes} or \texttt{No} token.

\begin{lstlisting}[style=prompt,basicstyle=\ttfamily\small]
1. mental health issues: <Yes|No>
2. physical health/characteristic: <Yes|No>
3. substance use: <Yes|No>
4. hopelessness: <Yes|No>
5. emotion dysregulation: <Yes|No>
6. low self-esteem: <Yes|No>
7. poor school performance: <Yes|No>
8. low socio-economic status: <Yes|No>
9. interpersonal violence: <Yes|No>
10. prior self-harm or suicidal thought/attempt: <Yes|No>
11. poor social support: <Yes|No>
12. interpersonal difficulty: <Yes|No>
13. dysfunctional family: <Yes|No>
14. exposure to others' suicide: <Yes|No>
15. stressful life event: <Yes|No>
16. traumatic experience: <Yes|No>
17. cognitive deficits: <Yes|No>
18. suicide means (with access): <Yes|No>
19. sexual orientation related issues: <Yes|No>
20. social support: <Yes|No>
21. coping strategy: <Yes|No>
22. psychological capital: <Yes|No>
23. sense of responsibility: <Yes|No>
24. meaning in life: <Yes|No>
\end{lstlisting}

The Task~2 zero-shot control used this same prompt and ordered answer grid without modification. It loaded the corresponding base instruction-tuned model without an adapter; thus, zero-shot denotes the absence of Task~2 parameter adaptation, not a different prompt or output format.

\subsection{GPT-5.4 Prompt-Ensemble Variants (Task~2)}

The three GPT-5.4 variants shared the user message and JSON response schema shown below. In the system-prompt listings, angle-bracketed definition blocks denote the factor-specific text inserted programmatically from the 24-factor codebook and, for the recommended variant, the mapped NVDRS coding guide. The surrounding instruction text is reproduced verbatim.

\begin{lstlisting}[style=prompt]
Post:
<POST>
\end{lstlisting}

The model returned only the following JSON object, with zero or more exact factor names:

\begin{lstlisting}[style=prompt]
{"factors": ["exact category name"]}
\end{lstlisting}

\subsubsection{Concise NVDRS-Essence Prompt}

\begin{pmdframed}[GPT-5.4 concise NVDRS-essence prompt]
\label{prompt:gpt54-essence}
\begin{lstlisting}[style=prompt,basicstyle=\ttfamily\scriptsize]
You are identifying suicide risk/protective factors from a social media post.

Task: choose all categories that are present in the post. Use only the exact category
names listed below. Return [] if no category is supported.

Category definitions:
<24 exact factor names and their concise definitions>

Decision rules:
- Select a category when it is directly stated or strongly implied by the post.
- Do not add broad categories from general distress alone.
- Do not output categories outside the list.
- Prefer precision for ambiguous categories, but include multiple categories when the
  post supports them.

Return ONLY valid JSON:
{"factors": ["exact category name"]}
\end{lstlisting}
\end{pmdframed}

\subsubsection{NVDRS-Recommended Prompt}

\begin{pmdframed}[GPT-5.4 NVDRS-recommended prompt]
\label{prompt:gpt54-recommended}
\begin{lstlisting}[style=prompt,basicstyle=\ttfamily\scriptsize]
You are identifying suicide risk/protective factors from a social media post.

Task: choose all categories that are present in the post. Use only the exact category
names listed below. Return [] if no category is supported.

The prompt has two styles:
1. Simple labels use concise category definitions.
2. Complex labels use longer NVDRS-style coding guidance and mapped NVDRS
   circumstance objects.

Simple category definitions:
<concise definitions for the 17 simple factors>

Complex NVDRS-style labels:
<coding guidance and mapped NVDRS circumstances for poor school performance,
interpersonal violence, prior self-harm or suicidal thought/attempt,
dysfunctional family, cognitive deficits, coping strategy, and meaning in life>

Decision rules:
- Select a category when it is directly stated or strongly implied by the post.
- For complex labels, a supported mapped NVDRS circumstance is sufficient evidence
  for the corresponding Task 2 factor, but do not force forensic/death-investigation
  details that are absent from social media text.
- Do not add broad categories from general distress alone.
- Do not output categories outside the list.
- Prefer precision for ambiguous categories, but include multiple categories when the
  post supports them.

Return ONLY valid JSON:
{"factors": ["exact category name"]}
\end{lstlisting}
\end{pmdframed}

\subsubsection{Factor-Specific Strictness-Control Prompt}

\begin{pmdframed}[GPT-5.4 factor-specific strictness-control prompt]
\label{prompt:gpt54-strict}
\begin{lstlisting}[style=prompt,basicstyle=\ttfamily\scriptsize]
You are identifying suicide risk/protective factors from a social media post.

Task: choose all categories that are present in the post. Use only the exact category
names listed below. Return [] if no category is supported.

Each category definition includes its own evidence threshold. Apply that threshold as
part of the definition for that specific category.

Category definitions:
<24 exact factor names and concise definitions, each followed by its assigned
strict, loose, or neutral evidence-threshold instruction>

Decision rules:
- Apply each label's evidence threshold; do not use one global threshold for all labels.
- Do not output categories outside the list.
- Include multiple categories when the post supports them.
- Return [] if no category is supported.

Return ONLY valid JSON:
{"factors": ["exact category name"]}
\end{lstlisting}
\end{pmdframed}

The three evidence-threshold instructions inserted into the strictness-control prompt were:

\begin{itemize}
    \item \emph{Strict:} Only code this factor when the post gives concrete evidence, such as a direct statement, a specific example, or an unmistakable description of the factor. Do not infer it from general distress, broad background context, or weak hints.
    \item \emph{Loose:} Code this factor when it is directly stated, strongly implied, or reasonably suggested by the context. If the evidence is plausible but not perfectly worded, err on the side of coding Yes.
    \item \emph{Neutral:} Code this factor when it is directly stated or strongly implied by the post. Use a balanced threshold: do not require exact wording, but do not infer it from weak evidence.
\end{itemize}

\end{document}